\documentclass[letterpaper]{article} 
\usepackage{aaai25}  
\usepackage{times}  
\usepackage{helvet}  
\usepackage{courier}  
\usepackage[hyphens]{url}  
\usepackage{graphicx} 
\usepackage{natbib}  
\usepackage{caption} 
\usepackage{algorithm}
\usepackage{algorithmic}

\usepackage{booktabs}
\usepackage{amssymb}
\usepackage{multirow} 
\usepackage{multicol}
\usepackage{algorithmic}
\usepackage{color,xcolor}
\usepackage{pifont}
\usepackage{adjustbox}
\usepackage{amsmath}

\usepackage{newfloat}
\usepackage{listings}
\DeclareCaptionStyle{ruled}{labelfont=normalfont,labelsep=colon,strut=off} 
\floatstyle{ruled}
\newfloat{listing}{tb}{lst}{}
\floatname{listing}{Listing}
\makeatletter
\def\thanks#1{\protected@xdef\@thanks{\@thanks
        \protect\footnotetext{#1}}}
\makeatother

\title{HSOD-BIT-V2: A New Challenging Benchmark \\ for Hyperspectral Salient Object Detection}
\author{
    Yuhao Qiu, Shuyan Bai, Tingfa Xu$^{\dagger}$, Peifu Liu, Haolin Qin, Jianan Li$^{\dagger}$\thanks{$\dagger$ Correspondence to: Tingfa Xu and Jianan Li.}
}
\affiliations{
    Beijing Institute of Technology

}

\usepackage{bibentry}

\begin{document}

\maketitle

\begin{abstract}
Salient Object Detection (SOD) is crucial in computer vision, yet RGB-based methods face limitations in challenging scenes, such as small objects and similar color features. Hyperspectral images provide a promising solution for more accurate Hyperspectral Salient Object Detection (HSOD) by abundant spectral information, while HSOD methods are hindered by the lack of extensive and available datasets. In this context, we introduce HSOD-BIT-V2, the largest and most challenging HSOD benchmark dataset to date. Five distinct challenges focusing on small objects and foreground-background similarity are designed to emphasize spectral advantages and real-world complexity. To tackle these challenges, we propose Hyper-HRNet, a high-resolution HSOD network. Hyper-HRNet effectively extracts, integrates, and preserves effective spectral information while reducing dimensionality by capturing the self-similar spectral features. Additionally, it conveys fine details and precisely locates object contours by incorporating comprehensive global information and detailed object saliency representations. Experimental analysis demonstrates that Hyper-HRNet outperforms existing models, especially in challenging scenarios. Dataset is avaliable at https://github.com/QYH-BIT/HSOD-BIT-V2.
\end{abstract}

%

\section{Introduction}

\begin{figure}[t]
    \centering  
    \includegraphics[width=\columnwidth]{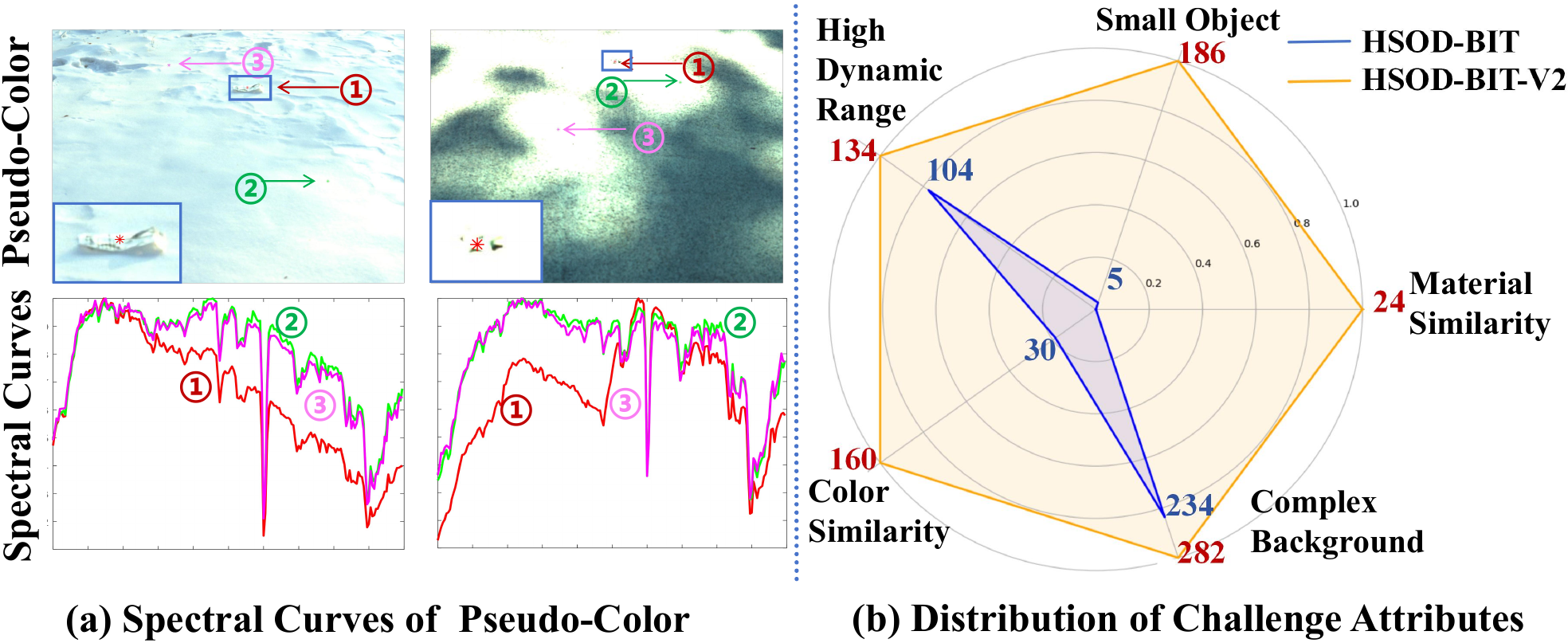}
    \caption{(a) Exemplary challenging scenarios from HSOD-BIT-V2, where objects are hard to identify in pseudo-color images but exhibit salience in spectral curves. (b) Challenge attributes of HSOD-BIT~\shortcite{paper45} and HSOD-BIT-V2, highlighting their capability to represent real-world challenges.} 
    \label{fig:motivation}
    \vspace{-2mm}
\end{figure}

Salient object detection (SOD) is crucial in various applications~\shortcite{paper2, paper5}, typically using RGB images to identify prominent objects.
However, RGB images struggle with accurate localization in challenging scenes, such as color similarity between foreground and background, due to reliance on shape and color features~\shortcite{paper9}. In contrast, spectral curves offer a more detailed characterization of objects' intrinsic properties~\shortcite{chenhuan, wangze}, as shown in Figure \ref{fig:motivation} (a).
Hyperspectral salient object detection (HSOD) methods utilize the abundant spectral information available in hyperspectral images (HSIs) to capture detailed object features, delivering enhanced performance even in challenging conditions~\shortcite{paper14}. Thus, integrating HSIs into SOD shows great promise for improving accuracy in challenging scenarios.

Convolutional neural networks (CNNs) enhance feature representation, boosting performance in HSOD task~\shortcite{paper16}. However, deep learning methods require extensive high-quality data, which is scarce in HSOD. 
Previous datasets primarily sourced from publicly available HSIs not curated for HSOD~\shortcite{paper35}, suffer from imprecise annotation and inadequate quantity and quality. Although the dedicated HS-SOD dataset~\shortcite{paper17} represents progress, it remains small and of low quality. The HSOD-BIT dataset~\shortcite{paper45} expands the dataset scale and introduces challenges like non-uniform lighting and overexposure, yet its limited diversity still hinders the full utilization of spectral advantages.

To bridge these gaps, we construct \textit{HSOD-BIT-V2}, the largest and most challenging HSOD dataset to date, featuring \textit{500 high-quality HSIs}. This dataset includes eight natural scene backgrounds and, for the first time in HSOD, introduces snowfields and fallen leaves, greatly enhancing diversity. The dataset emphasizes spectral advantages through five challenging attributes, focusing on small objects and foreground-background similarities. As depicted in Figure \ref{fig:motivation} (b), our dataset contains 417 challenging samples and outperforms HSOD-BIT across all attributes. With its expanded scale and varied challenges, HSOD-BIT-V2 provides enhanced data support for the HSOD task and serves as a new benchmark for algorithmic evaluation.

HSOD metods currently encounter three main challenges: (i) \textit{Spectral redundancy} raises computational costs, reduces effective information density, and diminishes detection accuracy due to the Hughes phenomenon~\shortcite{paper48}. Existing dimensionality reduction techniques, such as PCA~\shortcite{paper42}, often lead to information loss.
(ii) \textit{Effectively capturing spectral features} is vital due to the high spectral self-similarity and spatial sparsity of HSIs. Despite related research emphasizes this need~\shortcite{paper33}, fully exploiting spectral features remains difficult. (iii) \textit{Accurately distinguishing object contours} is essential for dense supervision tasks. Current methods often lose fine details through interpolation or pooling during downsampling~\shortcite{paper26}, making accurate contour detection challenging.

To tackle these challenges, we present Hyper-HRNet, a high-resolution HSOD network. Hyper-HRNet optimizes HSI utilization and minimizes spectral dimensionality while preserving crucial spectral information. It achieves this by synergizing CNN and Transformer for effective spectral feature extraction and reconstruction. Additionally, it supplements high-resolution flow decoding with intact global information and detailed object saliency representations to convey fine details and precisely locate object contours.

Firstly, Hyper-HRNet introduces \textit{Hyperspectral Attention Reconstruction} to effectively capture spectral features and address spectral redundancy. 
This component combines CNN and Transformer for effective spectral dimensionality reduction and reconstruction. The CNN adaptively captures high- and low-frequency spectral details, preserving edge and saliency features. Concurrently, the Transformer processes spectral feature map as a token to capture contextual spectral information and address long-range dependencies often inadequately handled by CNNs. 
This process harnesses the self-similar spectral features to enable seamless interactions in spectral-wise, thereby reducing spectral dimensionality and preserving effective spectral features.

Finally, Hyper-HRNet employs \textit{Global Ternary Perception Decoder} to convey fine details and precisely delineate object contours.
It fuses high-resolution flow from the backbone and enhances decoding through two modules: (i) \textit{Global Attention Feature Aggregator}, which utilizes features processed with PixelShuffle to produce a saliency map containing intact global information and offset fine details loss typically seen in multi-scale decoding; (ii) \textit{Ternary-Aware Weight}, which converts saliency predictions into ternary weights to emphasize essential regions between background and object, thereby improving contour localization accuracy.

Extensive experiments have been conducted to evaluate the performance of Hyper-HRNet on HSOD-BIT-V2, HSOD-BIT and HS-SOD datasets. Our model surpasses mainstream models, especially in challenging backgrounds.

Our contributions can be summarized as follows:
\begin{itemize}
\item We construct HSOD-BIT-V2, the largest and most challenging HSOD dataset to date, featuring five distinct attributes designed to highlight spectral advantages.
\item We introduce Hyper-HRNet, a novel network that effectively leverages HSIs to address spectral dimensionality and accurately delineate object contours.
\item We propose Hyperspectral Attention Reconstruction to optimize HSI utilization and reduce spectral dimensionality while preserving essential spectral information.
\item We develop Global Ternary Perception Decoder to enhance decoding by integrating intact global information and detailed object saliency representations.
\end{itemize}

\section{Related Work}

\noindent\textbf{Salient Object Detection}.
Traditional SOD methods relied on low-level features to measure saliency~\shortcite{paper15}, often emphasizing high-contrast edges rather than salient objects due to limited feature representation~\shortcite{paper21}. CNNs made great strides in SOD~\shortcite{paper24,paper25}.
Recent works adopt a two-stage framework to generate a trimap for ensuring clear edges~\shortcite{paper27}, while also enhancing global context modeling through Transformer-based patch-wise branchs~\shortcite{paper26}. 
Regrettably, these methods are limited to RGB data and tend to perform poorly when directly applied to HSIs.

\noindent\textbf{Hyperspectral Salient Object Detection}.
Despite advances in SOD, HSOD remains unexplored.  
Previous methods relied on shallow features like spectral gradients~\shortcite{paper28,paper30}, and utilized PCA for dimensionality reduction~\shortcite{paper29}, which often led to information loss or inadequate saliency capture. 
Deep learning models address these issues by incorporating spectral saliency and edge features to reduce information loss~\shortcite{paper42}, and using CNNs with knowledge distillation for dimensionality reduction~\shortcite{paper45}.
However, challenges remain in spectral feature utilization, information loss, and edge delineation. Therefore, we propose an attention-based component to better capture spectral self-similarity and a novel decoder to enhance object contours.

\noindent\textbf{Hyperspectral Salient Object Detection Datasets}. Acquiring HSIs is intricate, resulting in a scarcity of suitable data. 
Previous datasets, which relied on publicly available data not specifically curated for HSOD~\shortcite{paper36,paper37}, feature low-precision annotations and inferior quality.  
The first tailored HSOD dataset HS-SOD is small and limited to common scenes~\shortcite{paper17}. The HSOD-BIT dataset~\shortcite{paper45}, while larger and including some challenges, still lacks sufficient challenging data to fully showcase spectral advantages. Hence, HSOD requires larger, more diverse, and higher-quality datasets spanning various environmental scenarios.

\section{HSOD-BIT-V2 Dataset}
\subsection{Overview}
HSOD-BIT-V2 overcomes limitations in scale, quality, and challenge of current datasets. 
Table \ref{tab:statistical comparison} shows it surpassing existing HS-SOD and HSOD-BIT, with 500 HSIs,  $1240\times1680$ spatial resolutions, and 200 spectral bands. 
Unlike HS-SOD, which focuses on common scenes, and HSOD-BIT, with limited challenging data, HSOD-BIT-V2 covers 8 natural backgrounds with diverse and challenging data, highlighting small objects and foreground-background similarity.

\subsection{Dataset Construction}
\begin{figure}[t]
\centering
\includegraphics[width=\columnwidth]{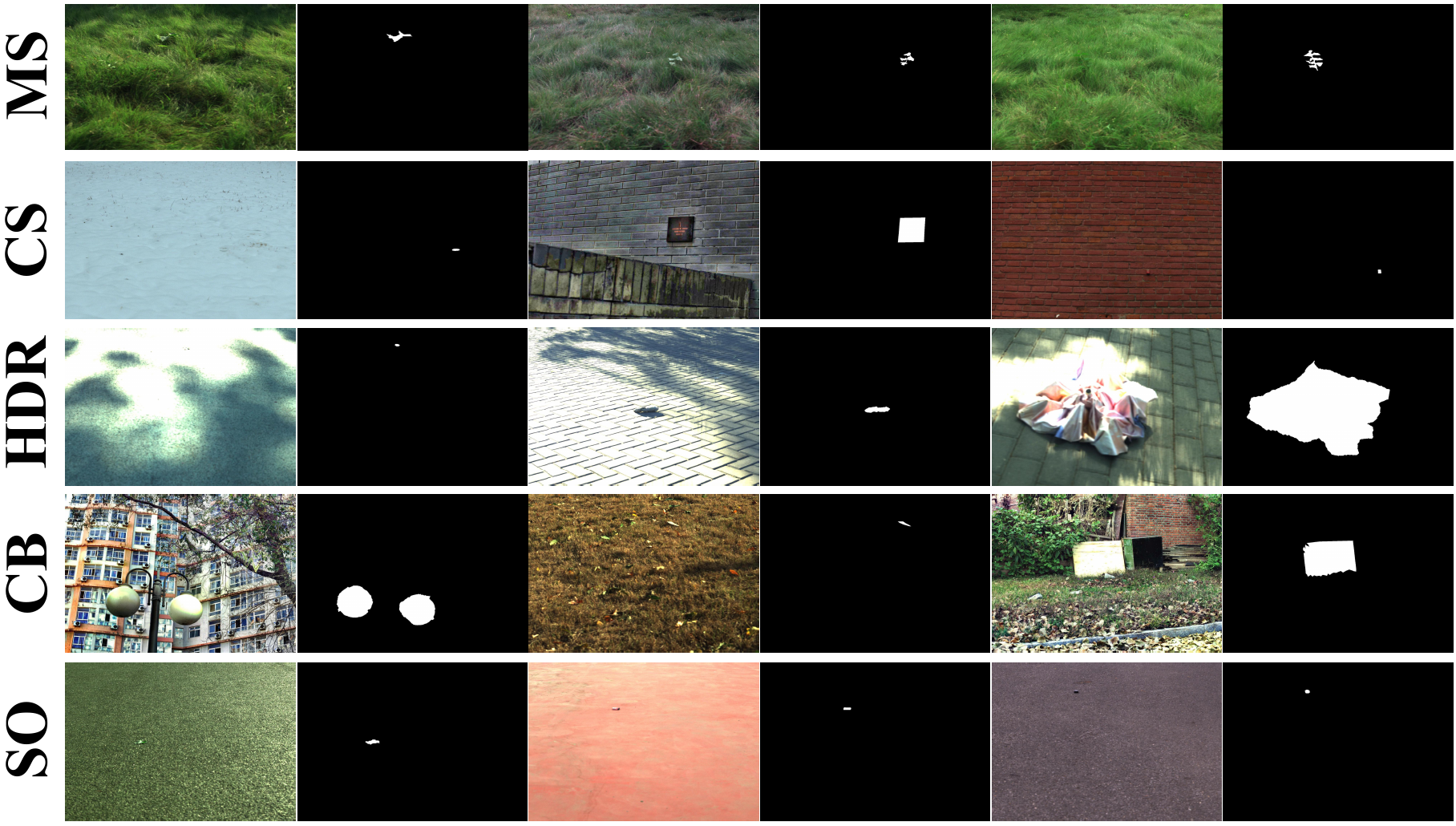}
\caption{Examples of pseudo-color images and corresponding ground truth from HSOD-BIT-V2.} 
\label{fig:dataset examples}
\vspace{-4mm}
\end{figure}

HSOD-BIT-V2 includes 8 natural backgrounds across various weather conditions, as shown in Figure \ref{fig:diagram of statistics}~(a), ensuring diversity and representativeness. 
Each scene type features multiple scenarios, with consistent imaging parameters for uniformity. 
To expand the dataset, we integrated and processed HSOD-BIT~\shortcite{paper45}, maintaining data coherence. 
Original data underwent dark current noise reduction, calibration, and quality evaluation, excluding low-quality or insufficiently challenging images. 
From the 500 processed data cubes, 406 images were used for training, and 94 for testing. 
Pseudo-color images were generated for easier annotation, with ground truth labels assigned using Matlab's ImageLabeler toolbox. Examples are shown in Figure \ref{fig:dataset examples}. 

\begin{table}[tp]
  \centering
  \renewcommand{\arraystretch}{0.4}
    \setlength{\tabcolsep}{2pt}
    \footnotesize
    \begin{tabular}{l|ccc}
    \toprule
    Property & \multicolumn{1}{c}{HS-SOD} & \multicolumn{1}{c}{HSOD-BIT} & \multicolumn{1}{c}{HSOD-BIT-V2} \\
    \midrule
    Data Volume & 60    & 319   & 500 \\
    Spatial Resolution & \multicolumn{1}{c}{$768\times 1024$} & \multicolumn{1}{c}{$1240\times 1680$} & \multicolumn{1}{c}{$1240\times 1680$} \\
    Spectral Bands & 81    & 200   & 200 \\
    Spectral Resolution &\multicolumn{1}{c}{5nm}  & \multicolumn{1}{c}{3nm} & \multicolumn{1}{c}{3nm} \\
    Spectral Range & \multicolumn{1}{c}{380-700nm} & \multicolumn{1}{c}{400-1000 nm} & \multicolumn{1}{c}{400-1000 nm} \\
    Challenges & 0     & 278   & 459 \\
    F-B similiarty & 0     & 30    & 160 \\
    Small object & 0     & 5     & 186 \\
    Scene Type & 4     & 6     & 8 \\
    \bottomrule
    \end{tabular}%
    \caption{Statistical Comparison of HSOD Datasets.}
  \label{tab:statistical comparison}%
  \vspace{-2mm}
\end{table}%

\subsection{Statistics}
We perform further rigorous statistical analysis on HSOD-BIT-V2 to validate its scientific integrity, providing a solid foundation for the splitting of training and testing sets.

\begin{figure}[t]
    \centering  
    \includegraphics[width=\columnwidth]{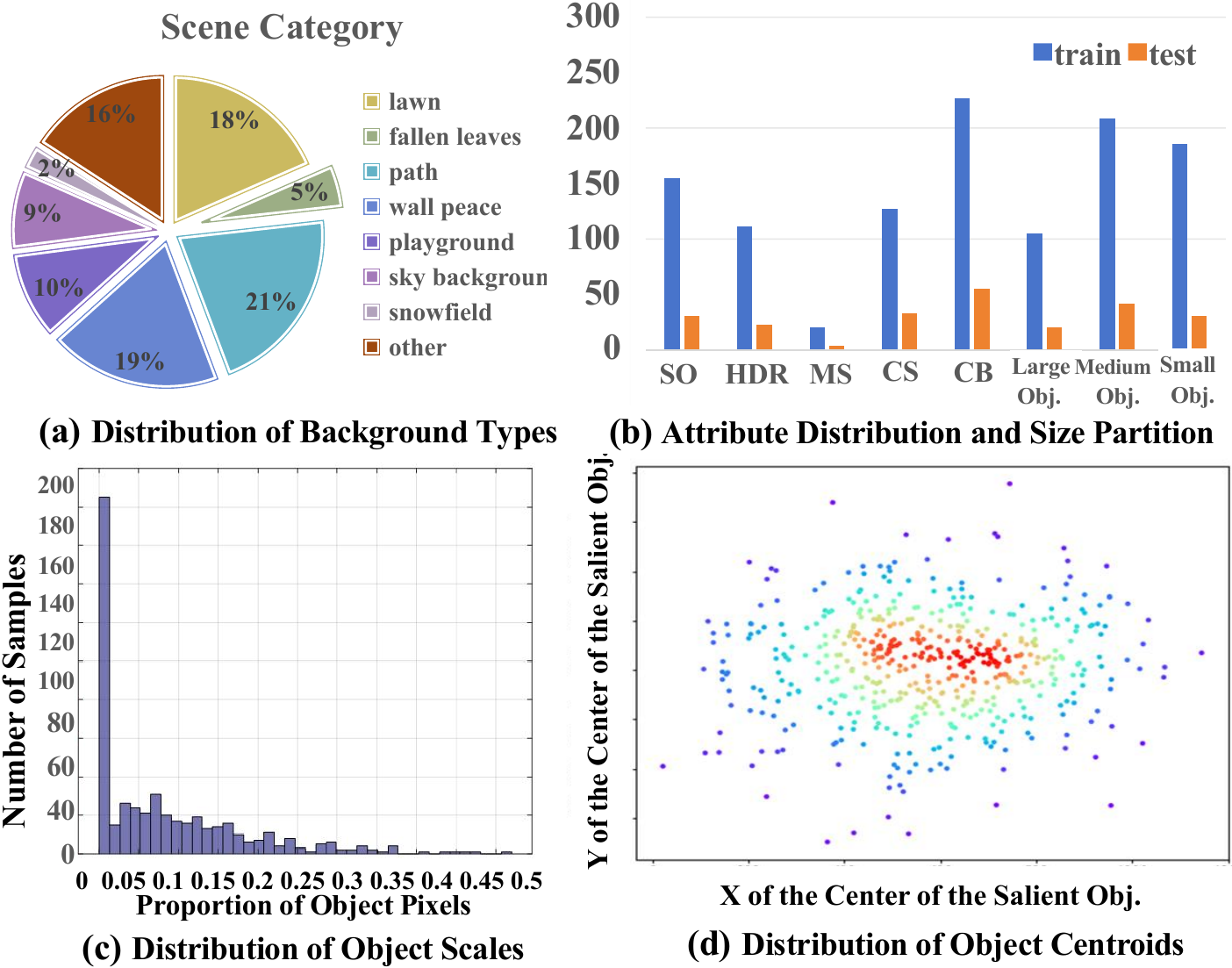}
    \caption{Diagram of HSOD-BIT-V2 statistics.} 
    \label{fig:diagram of statistics}
    \vspace{-4mm}
\end{figure}

\noindent\textbf{Challenge Attributes Statistics}. To evaluate HSOD method thoroughly, we categorize challenges into five attributes: \textit{Complex Background} (CB), \textit{Color Similarity} (CS), \textit{High Dynamic Range} (HDR), \textit{Small Object} (SO), and \textit{Material Similarity} (MS). MS is particularly difficult for HSI-based methods. Our dataset contains a substantial proportion of challenging data and notably numerous tiny objects. Figure \ref{fig:diagram of statistics}~(b) shows the distribution and sizes of these attributes, which are balanced to effectively address real-world challenges.

\noindent\textbf{Foreground Scale Analysis}. 
Our study of foreground scale shows a uniform distribution, with small objects (less than 1$\%$ of the image) comprising 38.4$\%$of the dataset, as shown in Figure \ref{fig:diagram of statistics}~(c). 
The diverse object scales improve the HSOD model's performance, making it more versatile and effective in detecting salient objects of varying sizes, which is crucial for real-world scenarios with objects at different distances.

\noindent\textbf{Centroid Spatial Distribution}. Figure \ref{fig:diagram of statistics}~(d) shows the spatial distribution of object centroids, represented by centroid probabilities across the dataset. 
Red regions denote dense clusters of centroid positions, with a uniform outward distribution from the center and higher concentration near the center. 
This pattern conforms to the natural inclination of human vision to prioritize prominent objects in the central field of view, while allocating less attention to the periphery.

\begin{figure*}[t]
\centering
\includegraphics[width=1.0\textwidth]{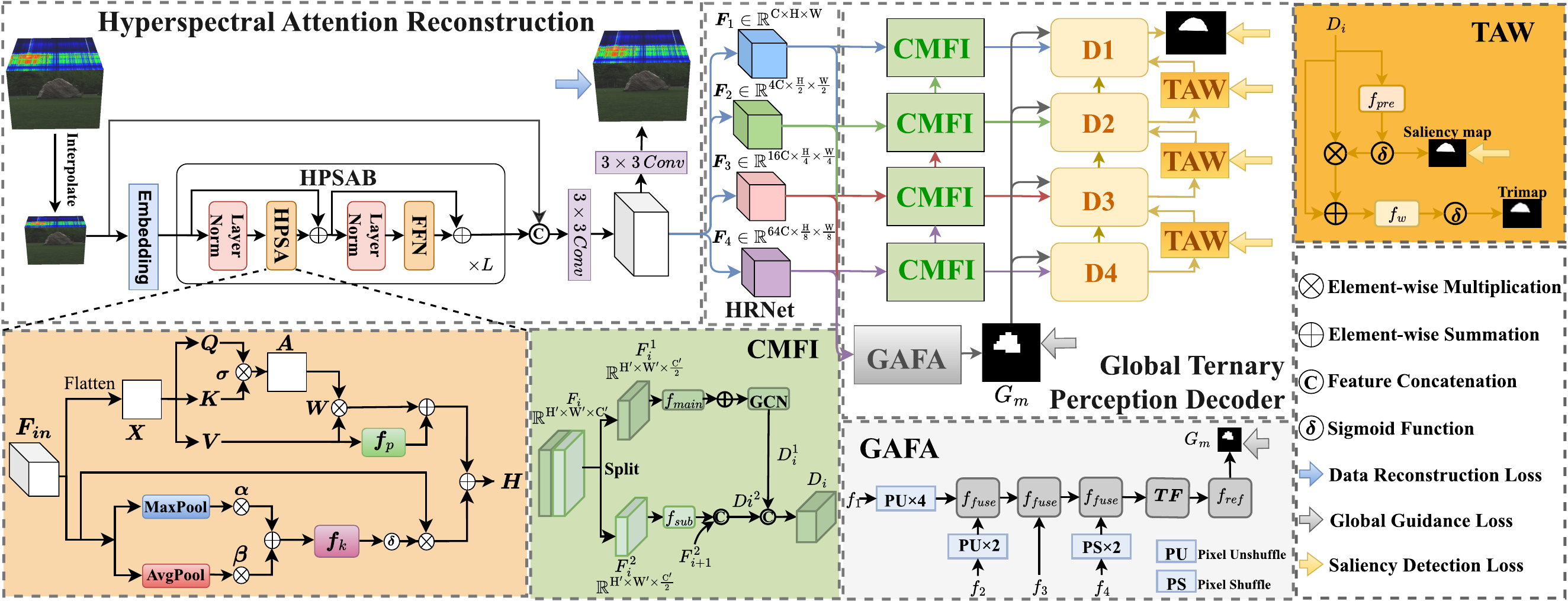}
\caption{The overall architecture of the proposed Hyper-HRNet is shown in the top part of the figure. The bottom part illustrates the detailed elucidation of composition within HPSAB and the blocks within GTPD.} 
\label{fig:Hyper-HRNet architecture}
\vspace{-4mm}
\end{figure*}

\section{Method}
Given a HSI $\boldsymbol{I} \in \mathbb{R}^{\mathrm{H} \times \mathrm{W} \times \mathrm{C}}$, HSOD aims to generate a saliency map $\boldsymbol{Y} \in \mathbb{R}^{\mathrm{H} \times \mathrm{W} \times \mathrm{1}}$, a binary image highlighting salient object. Hyper-HRNet interpolates and reconstructs HSI to preserve crucial information by capturing spectral self-similarity. It also retains high-resolution flow and integrates intact global information and detailed object saliency to enhance decoding results, depicted in Figure \ref{fig:Hyper-HRNet architecture}.

\subsection{Hyperspectral Attention Reconstruction}
Hyper-HRNet utilizes the Hyperspectral Attention Reconstruction (HAR) to downsample the channels from $C$ to $C'$ by interpolating every 4 channels into 1 from $\boldsymbol{I}$, then reconstructing the dimension-reduced HSI, depicted in Figure \ref{fig:Hyper-HRNet architecture}.
HAR first applies a $3 \times 3$ convolution for embedding, then uses cascaded Hybrid Perceptual Spectral Attention Reconstruction Blocks (HPSAB). These blocks incorporate Transformer-like architecture with Hybrid Perceptual Spectral Attention (HPSA) which combines Transformer-based Multi-head Spectral-wise Self-Attention (MSSA) and CNN-based Adaptive Spectral Attention Mechanism (ASAM) to capture spectral-wise self-similar relationships. 
HAR effectively reconstructs the interpolated data, addressing spectral redundancy while preserving essential information.

\noindent\textbf{Multi-head Spectral-wise Self-Attention (MSSA).} MSSA enhances the self-attention mechanism to capture spectral-wise contextual relationships.
Given the input $\boldsymbol{F}_{in}\in \mathbb{R}^{\mathrm{H} \times \mathrm{W} \times \mathrm{C'}}$ obtained through interpolation and embedding from $\boldsymbol{I}$, it is reshaped into tokens $\boldsymbol{X} \in \mathbb{R}^{\mathrm{HW} \times \mathrm{C'}}$ and linearly projected into $\boldsymbol{Q},\boldsymbol{K},\boldsymbol{V}\in \mathbb{R}^{\mathrm{HW} \times \mathrm{C'}}$. Then, $\boldsymbol{Q}$, $\boldsymbol{K}$, and $\boldsymbol{V}$ are split into $\rm{N}$ parts along the spectral channel dimension: $\boldsymbol{Q} = \left [ \boldsymbol{Q}_{1},\cdots ,\boldsymbol{Q}_{\rm{N}}  \right ]$, $\boldsymbol{K} = \left [ \boldsymbol{K}_{1},\cdots , \boldsymbol{K}_{\rm{N}}  \right ]$, and $\boldsymbol{V} = \left [ \boldsymbol{V}_{1},\cdots ,\boldsymbol{V}_{\rm{N}}  \right ]$. Then, MSSA treats each spectral representation as a token and computes self-attention $\boldsymbol{SSA}_{j}$: 
\begin{equation}
\begin{aligned}
\boldsymbol{A}_{j} &= \text{softmax}(\boldsymbol{\sigma} _{j}\boldsymbol{K}_{j}^{T}\boldsymbol{Q}_{j}), \quad
\boldsymbol{SSA}_{j} = \boldsymbol{V}_{j}\boldsymbol{A}_{j},
\end{aligned}
\end{equation}
where $\boldsymbol{K}_{j}^{T}$ denotes the transpose of $\boldsymbol{K}_{j}$. Due to the significant variation in spectral density across wavelengths, we reweight the matrix multiplication $\boldsymbol{K}_{j}^{T}Q_{j}$  within $\boldsymbol{A}_{j}$ using a learnable parameter $\boldsymbol{\boldsymbol{\sigma}} _{j} \in \mathbb{R}^{\mathrm{1}}$ to adapt to the spectral density variation. The outputs from $\mathrm{N}$ heads are then concatenated, linearly projected, and position embedding to generate the output feature maps $\boldsymbol{M} \in \mathbb{R}^{\mathrm{H} \times \mathrm{W} \times \mathrm{C'}}$:
\begin{equation}
\label{11}
\boldsymbol{M} = (\sum\limits_{j=1}^{\mathrm{N}} (\boldsymbol{SSA}_{j}))\boldsymbol{W} + \boldsymbol{f}_{p}(\boldsymbol{V}), 
\end{equation}
where $\boldsymbol{W}\in\mathbb{R}^{\mathrm{C'}\times \mathrm{C'}}$ is a learnable parameter, $\boldsymbol{f}_{p}(\cdot )$ is position embedding function including two depth-wise $3\times3$ convolutions, GELU activation, and reshape operation.

\noindent\textbf{Adaptive Spectral Attention Mechanism (ASAM).} To adaptively extract spectral details, ASAM utilizes the high-frequency saliency feature from max-pooling and the low-frequency degree feature from average-pooling~\shortcite{paper33}.
The input feature $\boldsymbol{F}_{in}$ is processed through two branches along spectral dimension: max-pooling for discriminative object features $\boldsymbol{F}_{max}$, and average-pooling for holistic object features $\boldsymbol{F}_{avg}$. Because varying emphasis across different stages, learnable parameters $\boldsymbol{\alpha}$ and $\boldsymbol{\beta}$ are used to weigh $\boldsymbol{F}_{avg}$ and $\boldsymbol{F}_{max}$. The weighted tensors are combined to produce the adaptive spectral feature $\boldsymbol{F}_{add}\in \mathbb{R}^{1\times1\times \mathrm{C'}}$:
\begin{equation}
\boldsymbol{F}_{add} = \frac{1}{2}(\boldsymbol{F}_{avg}\oplus \boldsymbol{F}_{max})\oplus\boldsymbol{\alpha} \otimes \boldsymbol{F}_{avg}\oplus\boldsymbol{\beta}  \otimes \boldsymbol{F}_{max},  
\end{equation}
where $\otimes$ means element-wise multiplication, $\oplus$ means element-wise summation. After applying the Sigmoid activation and performing element-wise multiplication with $\boldsymbol{F_{in}}$, the output feature maps $\boldsymbol{S}$ are obtained as follows:
\begin{equation}
\boldsymbol{S} = \boldsymbol{F}_{in}\times\boldsymbol{\delta }  (\boldsymbol{f}_{k} (\boldsymbol{F}_{add})),
\end{equation}
where $\boldsymbol{\delta } $ stands for Sigmoid activation function, $\boldsymbol{f}_{k}$ stands for 1D with an adaptive kernel size of k~\shortcite{paper33}. The final Spectral-wise Attention feature $\boldsymbol{H}\in\mathbb{R}^{\mathrm{H} \times \mathrm{W} \times \mathrm{C'}}$ is obtained by element-wise summing $\boldsymbol{S}$ and $\boldsymbol{M}$.

Finally, the reconstructed image is restored to $C$ channel using a $3\times3$ convolution. These processes are supervised by $\boldsymbol{I}$ to maximize the preservation of spectral information.

\subsection{Global Ternary Perception Decoder}
Hyper-HRNet enhances object contours and decoding results through Global Ternary Perception Decoder (GTPD). 
Fusing cross-scale high-resolution flow from HRNet backbone~\shortcite{paper18}, GTPD supplements intact global information and ternary contour-aware saliency for precise decoding and accurate saliency predictions, as shown in Figure \ref{fig:Hyper-HRNet architecture}.

\noindent\textbf{Cross-level Multi-scale Feature Interaction (CMFI)}. To discern scale and positional changes of objects in multi-scale features and highlight salient regions, GTPD uses CMFI based on the Split-Transform-Merge strategy~\shortcite{paper38}. The multi-scale features $\boldsymbol{F} = \{\boldsymbol{F}_i \in \mathbb{R}^{\mathrm{C_i} \times \mathrm{H_i} \times \mathrm{W_i}}|i=1,2,3,4\}$ from HRNet are split along the channel dimension into two parts for CMFI: $\left \{\boldsymbol{F}_{i}^{1}, \boldsymbol{F}_{i}^{2} \right \}$. $F_{i}^{1}$ captures local context at small scales and expands the receptive field through GCN~\shortcite{paper39}, while $\boldsymbol{F}_{i}^{2}$ facilitates cross-scale interaction using its rich shallow-level details and semantic information in deep-level features $\boldsymbol{F}_{i+1}^{2}$.
They are treated as follows:
\begin{equation} 
\boldsymbol{D}_{i}^{1} = \boldsymbol{f}_{GCN}(\boldsymbol{f}_{main} (\boldsymbol{F}_{i}^{1})+\boldsymbol{F}_{i}^{1}), 
\end{equation}
\begin{equation} 
\boldsymbol{D}_{i}^{2} = \text{Cat}(\boldsymbol{f}_{sub}(\boldsymbol{F}_{i}^{2}),\boldsymbol{F}_{i+1}^{2}),
\end{equation}
where $\boldsymbol{f}_{main}(\cdot )$ performs downsampling via average pooling and $3\times3$ convolution, and corresponding upsampling. $\boldsymbol{f}_{GCN}(\cdot )$ denotes GCN. $\boldsymbol{f}_{sub}(\cdot )$ includes one $1\times1$ and two $3\times3$ convolutions. $\text{Cat}(\cdot )$ denotes concatenation. Finally, the decoded output features $\boldsymbol{D}_{i}$ are obtained as:
\begin{equation} 
\boldsymbol{D}_{i} = \text{Cat}(\boldsymbol{D}_{i}^{1},\boldsymbol{D}_{i}^{2}).
\end{equation}

\noindent\textbf{Global Attention Feature Aggregator (GAFA)}. To offset the loss of fine details and and limitations in capturing long-range dependencies in CNN-based multi-scale decoding, GTPD uses GAFA to incorporate intact global information into the decoding process. GAFA utilizes intact contextual features to generate global saliency via Transformer under supervision. Specifically, $\boldsymbol{F}_{i}$ employs Pixel Shuffle into spatial dimensions of $20\times20$ as $\boldsymbol{f}_{i}$ which is then concatenated. After fusing $\boldsymbol{f}_{i}$, GAFA generates the global saliency map $\boldsymbol{G}_{m}$ via Transformer and linear mapping  as follows:
\begin{equation}
\boldsymbol{G}_{m} = \boldsymbol{\delta } (\boldsymbol{f}_{ref} (\boldsymbol{TF}(\sum_{i =1}^{n} \boldsymbol{f}_{fuse}(\boldsymbol{f}_{i}))),
\end{equation}
where $\boldsymbol{f}_{fuse}(\cdot )$ applies $3\times3$ convolution, Batch Normalization, ReLU activation, and reshape operation which transforms the shape from $\mathbb{R}^{\mathrm{C_i} \times \mathrm{H_i} \times \mathrm{W_i}}$ to $\mathbb{R}^{\mathrm{H_i}\mathrm{W_i}\times \mathrm{C_i}}$. $\boldsymbol{f}_{ref}$ is realized via MLP. $\boldsymbol{TF}$ involves the original self-attention as:
\begin{equation}
\boldsymbol{TF}(\boldsymbol{Q},\boldsymbol{K},\boldsymbol{V}) = \text{softmax}(\frac{\boldsymbol{Q}\boldsymbol{K}^{T}}{\sqrt{d_{h}}})\boldsymbol{V}.
\end{equation}
To obtain the ground truth $\boldsymbol{GT}_{m}$ for $\boldsymbol{G}_{m}$, ground truth map employs Pixel Unshuffle into spatial dimensions of $20\times20$, preserving complete information. $\boldsymbol{GT}_{m}$ is obtained as:
\begin{equation}
\boldsymbol{GT}_{m} = {\text{maxc}} (\text{PS}(\boldsymbol{G}_{m})),
\end{equation}
where $\text{PS}(\cdot )$ denotes the Pixel Unshuffle operation, and ${\text{maxc}}(\cdot )$ signifies maximum along the channels.

\noindent \textbf{Ternary-Aware Weight (TAW)}. To enhance object contour delineation, GTPD leverages TAW to generate ternary-aware saliency layer by layer, focusing on uncertain region. Saliency prediction is categorized into three regions: object (saliency around 1), background (saliency around 0), and uncertain region(contours between object and background). The uncertain region is crucial in challenging scenarios. TAW first utilizes the decoded features $\boldsymbol{D}_{i}$ to produce the saliency prediction $\boldsymbol{P}_{i}$ as:

\begin{equation}
\boldsymbol{P}_{i} = \boldsymbol{\delta }  (\boldsymbol{f}_{pre} (\boldsymbol{D}_{i})),
\end{equation}
where $\boldsymbol{f}_{pre}(\cdot )$ uses $3\times3$ convolution.
The saliency prediction from the lower layer generates ternary contour-aware saliency, producing a Trimap $\boldsymbol{T}_{i}$ as weights for subsequent decoding stages. $\boldsymbol{T}_{i}$ labels each pixel: 1 for object, 0 for background, and 2 for uncertain regions. The TAW process can be summarized as follows:
\begin{equation}
\boldsymbol{T}_{i} = \text{softmax}(\boldsymbol{f}_{w} ((\boldsymbol{D}_{i} \otimes \boldsymbol{P}_{i})+\boldsymbol{D}_{i})),
\end{equation}
where $\boldsymbol{f}_{w}(\cdot )$ fuction employs a $3\times3$ convolution.

Ultimately, hierarchical decoding is optimized using the supplementary features $\boldsymbol{G}_{m}$ and $\boldsymbol{T}_{i}$, with dense supervision. The output from the topmost layer is the final saliency map.

\subsection{Loss Function}
Hyper-HRNet is trained with a hybrid loss function that comprises data reconstruction loss $\boldsymbol{L}_{s}$, saliency detection loss $\boldsymbol{L}_{sod}$, and global guidance loss $\boldsymbol{L}_{g}$, defined as follows:
\begin{equation}
\boldsymbol{L}_{m} = \boldsymbol{L}_{s} + \boldsymbol{L}_{sod} +\boldsymbol{L}_{g}.
\end{equation}

$\boldsymbol{L}_{s}$ evaluates the discrepancy between the restored image and the original data.
$\boldsymbol{L}_{sod}$ measures the deviation between the predicted saliency map and the ground truth.
$\boldsymbol{L}_{g}$ supervises global saliency map by its corresponding ground truth.

\section{Experiment}
We evaluate Hyper-HRNet on the HSOD-BIT-V2, HSOD-BIT, and HS-SOD datasets. To ensure fairness, all comparison methods are independently trained and tested on the same conditions across three datasets. Further experiments and details are available in the supplementary material.

\begin{table*}[!t]
  \centering \small
  \renewcommand{\arraystretch}{0.4}
  \setlength{\tabcolsep}{0.9pt}
    \footnotesize
  \begin{tabular}{l|cccccc|cccccc|cc}
    \toprule
          \multicolumn{1}{p{6.625em}|}{Dataset} & \multicolumn{6}{c|}{HSOD-BIT-V2}  & \multicolumn{6}{c|}{HSOD-BIT} & \multicolumn{2}{c}{}\\
    \midrule
          \multicolumn{1}{p{6.625em}|}{Metrics} & $MAE\downarrow$   & $PRE\uparrow$   & $REC\uparrow$   & $avgF_{1} \uparrow$  & $AUC\uparrow$   & $CC\uparrow$   & $MAE\downarrow$   & $PRE\uparrow$   & $REC\uparrow$   & $avgF_{1} \uparrow$  & $AUC\uparrow$   & $CC\uparrow$  & \#Params & FLOPs\\
    \midrule
    \multicolumn{15}{c}{\textit{RGB-based SOD Methods}} \\
    \midrule
    
    Itti~\shortcite{paper15} & 0.230  & 0.280  & 0.419  & 0.240  & 0.803  & 0.277  & 0.252  & 0.335  & 0.399  & 0.341  & 0.793  & 0.351  & -  & - \\
          BASNet~\shortcite{paper44} & 0.049  & 0.638  & 0.634  & 0.553  & 0.876  & 0.618  & 0.071  & 0.741  & 0.742  & 0.695  & 0.901  & 0.703  &  87.06 M & 127.56 G\\
          U2Net~\shortcite{paper43} & 0.046  & 0.649  & 0.597  & 0.513  & 0.948  & 0.621  & 0.062  & 0.814  & 0.683  & 0.739  & 0.951  & 0.746  &  44.01 M  & 47.65 G\\
          SelfReformer~\shortcite{paper26} & 0.048  & 0.581  & 0.498  & 0.528  & 0.827  & 0.530  & 0.068  & 0.766  & 0.628  & 0.704  & 0.884  & 0.676  & 90.70 M  & 128.26 G\\
    \midrule
    \multicolumn{15}{c}{\textit{HSI-based HSOD Methods}} \\
    \midrule
    
    SAD~\shortcite{paper28} & 0.177  & 0.335  & 0.398  & 0.253  & 0.863  & 0.331  & 0.209  & 0.395  & 0.350  & 0.364  & 0.822  & 0.395  & -  & -\\
          SED~\shortcite{paper28} & 0.106  & 0.359  & 0.178  & 0.237  & 0.781  & 0.264  & 0.138  & 0.415  & 0.131  & 0.345  & 0.746  & 0.301  & -  & - \\
          SG~\shortcite{paper28} & 0.168  & 0.342  & 0.350  & 0.243  & 0.823  & 0.298  & 0.188  & 0.401  & 0.278  & 0.351  & 0.782  & 0.363  & -  & - \\
          SED-SAD~\shortcite{paper28} & 0.180  & 0.345  & 0.367  & 0.265  & 0.865  & 0.333  & 0.208  & 0.400  & 0.317  & 0.381  & 0.828  & 0.407  & -  & - \\
          SED-SG~\shortcite{paper28} & 0.165  & 0.332  & 0.302  & 0.243  & 0.820  & 0.287  & 0.189  & 0.391  & 0.247  & 0.381  & 0.776  & 0.351  & -  & - \\
          SUDF~\shortcite{paper16} & 0.166  & 0.375  & 0.614  & 0.362  & 0.873  & 0.412  & 0.203  & 0.545  & 0.619  & 0.528  & 0.910  & 0.582  &  0.10 M & 82.90 G\\
          SMN~\shortcite{paper42} & 0.039  & 0.607  & 0.713  & 0.575  & 0.915  & 0.639  & 0.034  & 0.837  & 0.868  & 0.751  & 0.963  & 0.846  &  10.23 M  & 14.76 G\\
          DMSSN~\shortcite{paper45} & 0.072  & 0.635  & 0.602  & 0.548  & 0.830  & 0.553  & 0.086  & 0.663  & 0.637  & 0.637  & 0.852  & 0.625  & 1.76 M  & 10.89 G\\
    \midrule
          Hyper-HRNet & \textbf{0.028}  & \textbf{0.653}  & \textbf{0.764}  & \textbf{0.589}  & \textbf{0.988}  & \textbf{0.737}  & \textbf{0.020}  & \textbf{0.854}  & \textbf{0.891}  & \textbf{0.795}  & \textbf{0.996}  & \textbf{0.916}  & 29.57 M  & 18.96 G\\
           Hyper-HRNet-Lite & 0.046  & 0.590  & 0.689  & 0.550  & 0.940  & 0.641  & 0.026  & 0.845  & 0.878  & 0.770  & 0.987  & 0.907  & 7.24 M  & 7.85 G\\
    \bottomrule
    \end{tabular}%
    \caption{Quantitative Results on HSOD-BIT and HSOD-BIT-V2 Datasets.}
  \label{tab:HSOD-BIT quantitative}%
\end{table*}%

\subsection{Results On HSOD-BIT-V2 and HSOD-BIT}
\noindent\textbf{Quantitative Analysis}. Table \ref{tab:HSOD-BIT quantitative} provides a quantitative comparison of Hyper-HRNet with existing methods on HSOD-BIT-V2 and HSOD-BIT datasets. The results show that our method outperforms both RGB- and HSI-based methods across all metrics. Notably, it surpasses the soTA HSI-based method SMN and RGB-based method U2Net by 0.051 and 0.266 in REC, 0.098 and 0.207 in CC, and 0.073 and 0.040 in AUC on HSOD-BIT-V2. These gains highlight the effectiveness of our approach. While RGB-based methods perform well on simpler samples, they struggle in more challenging scenarios, emphasizing the limitations of converting HSI to pseudo-color images for RGB-based SOD methods.

Furthermore, the analysis reveals that performance on HSOD-BIT-V2 is significantly lower than on HSOD-BIT, highlighting the increased challenges presented by our dataset. Traditional HSI-based methods exhibit variable performance, likely due to the enhanced denoising in HSOD-BIT-V2, which improves spectral quality.

\begin{figure}[t]
\centering
\includegraphics[width=\columnwidth]{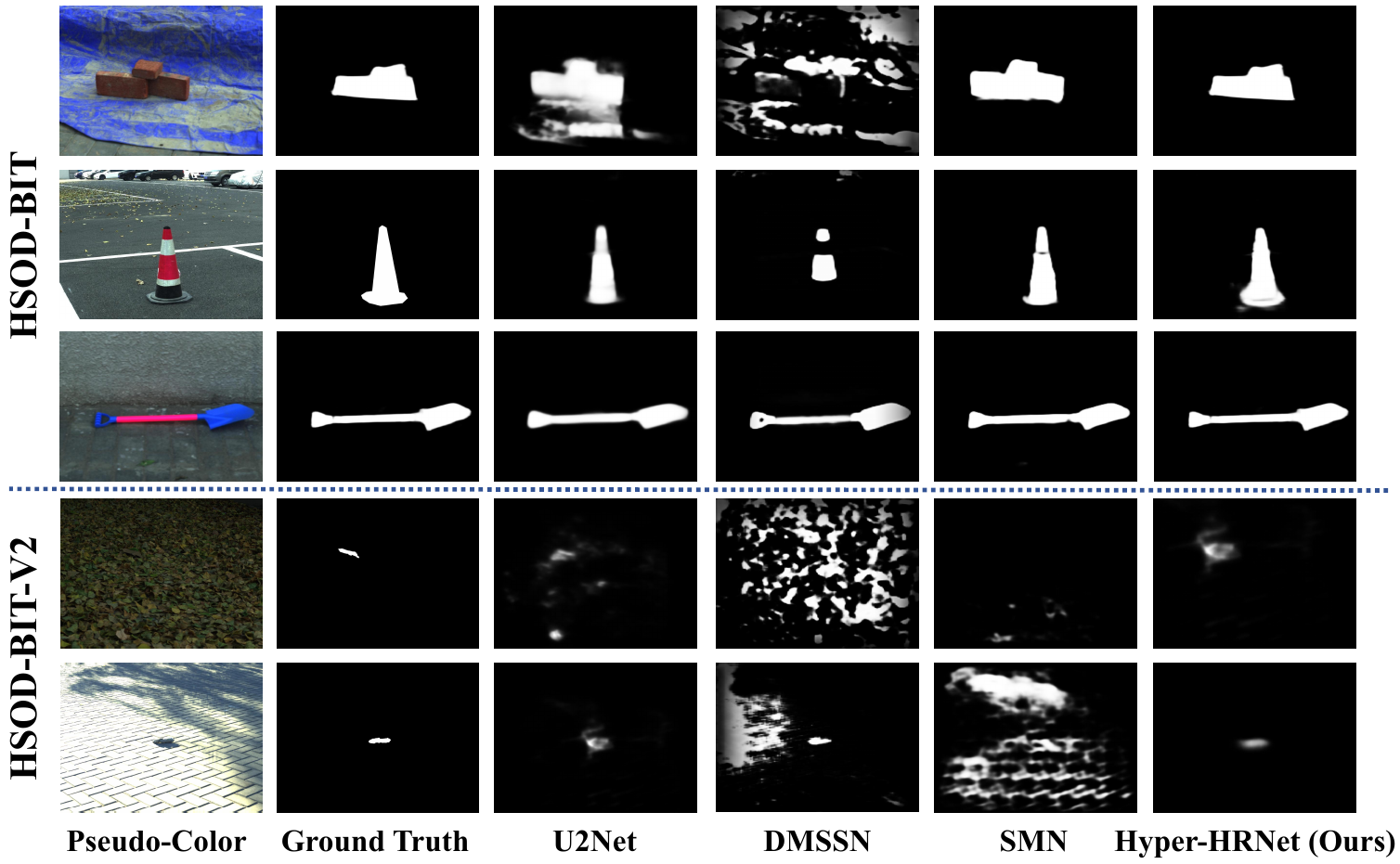}
\caption{Qualitative results on HSOD-BIT-V2 and HSOD-BIT datasets. Hyper-HRNet has best detection performance.}
\label{fig:HSOD-BIT qualitative}
\end{figure}

\begin{table}[t]
    \centering
    \setlength{\tabcolsep}{0.1pt}
    \scriptsize
    \begin{tabular}{l|cc|cc|cc|cc|cc}
      \toprule
      \multicolumn{1}{p{6em}|}{\textbf{Challenges}} & \multicolumn{2}{c|}{\textbf{CB}} & \multicolumn{2}{c|}{\textbf{SC}} & \multicolumn{2}{c|}{\textbf{HDR}} & \multicolumn{2}{c|}{\textbf{SO}} & \multicolumn{2}{c}{\textbf{SM}} \\
      \midrule
      \multicolumn{1}{p{7.055em}|}{Metrics} & MAE↓  & AUC↑  & MAE↓  & AUC↑ & MAE↓  & AUC↑  & MAE↓  & AUC↑  & MAE↓  & AUC↑ \\
      \midrule
      U2Net & 0.054  & 0.949  & 0.058  & 0.967  & 0.047  & 0.919  & 0.012  & 0.964  & \textbf{0.020 } & 0.987  \\
      SelfReformer & 0.058  & 0.841  & 0.066  & 0.753  & 0.049  & 0.818  & 0.019  & 0.720  & 0.027  & 0.837  \\
      SMN & 0.041  & 0.897  & 0.030  & 0.851  & 0.057  & 0.923  & 0.036  & 0.809  & 0.042  & 0.857  \\
      DMSSN & 0.089  & 0.774  & 0.068  & 0.782  & 0.050  & 0.846  & 0.060  & 0.669  & 0.095  & 0.502  \\
      \midrule
      \textbf{Hyper-HRNet} & \textbf{0.029 } & \textbf{0.979 } & \textbf{0.014 } & \textbf{0.973 } & \textbf{0.018 } & \textbf{0.989 } & \textbf{0.008 } & \textbf{0.979 } & 0.030  & \textbf{0.994 } \\
      \bottomrule
    \end{tabular}
    \caption{Attribute Evaluations on HSOD-BIT-V2 Dataset.}
    \label{tab:challenging attributes}
    \vspace{-2mm}
\end{table}

\noindent\textbf{Qualitative Analysis}. Figure \ref{fig:HSOD-BIT qualitative} presents visual comparisons of saliency maps generated by Hyper-HRNet on HSOD-BIT-V2 and HSOD-BIT datasets, alongside several existing HSOD methods.  Hyper-HRNet outperforms other methods by leveraging spectral information to minimize background noise and enhance object localization in challenging scenes. Moreover, by preserving essential spectral details and integrating global and key region information during decoding, Hyper-HRNet produces sharper contours.

\noindent\textbf{Attribute-based Evaluations}. We evaluate our approach on five challenging attributes of HSOD-BIT-V2, as detailed in Table \ref{tab:challenging attributes}. Our method outperforms both RGB- and HSI-based methods across most attributes. In MS attributes, where foreground and background consist of chemically similar materials and closely matching colors, extracting discriminative features from HSIs is more challenging than from RGB images. Nonetheless, Hyper-HRNet consistently exceeds other HSI-based methods and nearly matches the performance of the soTA RGB-based methods U2Net, improving AUC by 0.005 and falling short of MAE by only 0.010. Specifically, Hyper-HRNet outperforms the soTA HSI-based approach SMN, with a 0.137 improvement in AUC and a 0.012 reduction in MAE for MS attributes. Additionally, RGB-based methods perform poorly on other challenging attributes, underscoring the limitations of applying SOD methods to pseudo-color images derived from HSIs.

\begin{figure}[t]
\centering
\includegraphics[width=0.95\linewidth]{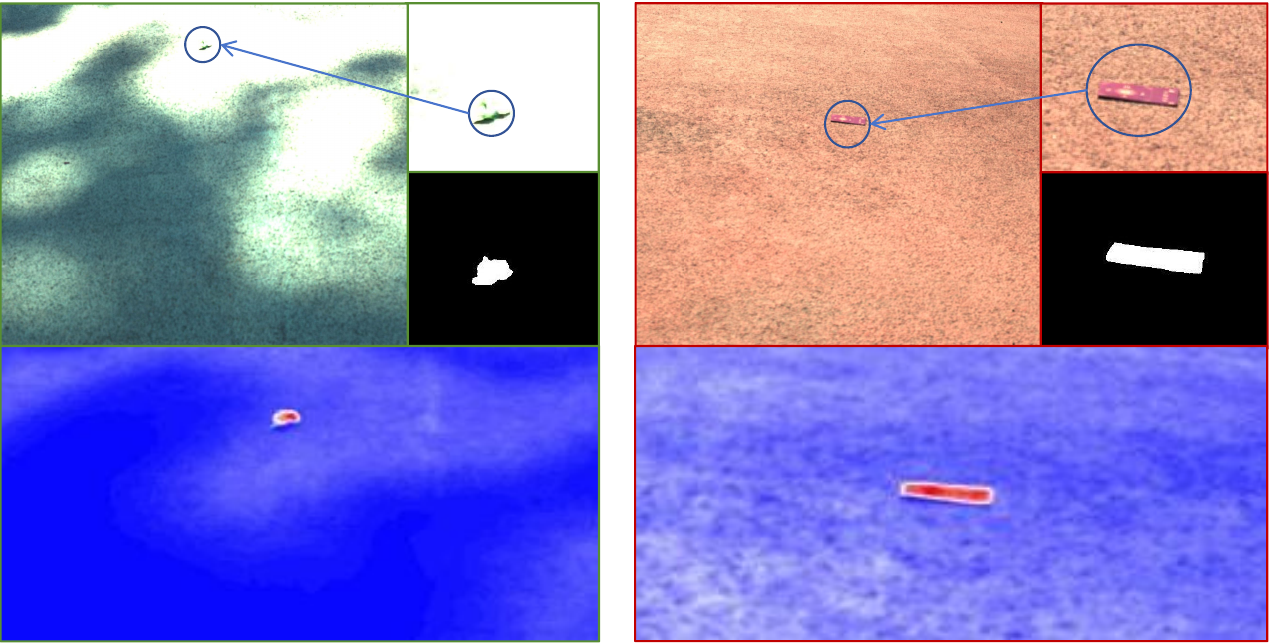}
\caption{Visualization of Hybrid Perceptual Spectral Attention features by HARM block. Attention features effectively preserve the salient information of the salient objects.} 
\label{fig:hotmap}
\vspace{-4mm}
\end{figure}

\begin{table}[t]
  \centering
    \renewcommand{\arraystretch}{0.6}
    \footnotesize
    \begin{tabular}{l|rrrr}
    \toprule
    \multicolumn{1}{p{6.025em}|}{Methods} & \multicolumn{1}{c}{$MAE\downarrow$} & \multicolumn{1}{c}{$avgF_{1} \uparrow$} & \multicolumn{1}{c}{$AUC\uparrow$} & \multicolumn{1}{c}{$CC\uparrow$} \\
    \midrule
    Itti~\shortcite{paper15} & 0.246  & 0.237  & 0.783  & 0.268  \\
    SAD~\shortcite{paper28} & 0.236  & 0.235  & 0.834  & 0.295  \\
    SED~\shortcite{paper28} & 0.185  & 0.236  & 0.817  & 0.277  \\
    SG~\shortcite{paper28} & 0.218  & 0.233  & 0.827  & 0.296  \\
    SED-SAD~\shortcite{paper28} & 0.209  & 0.250  & 0.830  & 0.286  \\
    SED-SG~\shortcite{paper28} & 0.188  & 0.240  & 0.826  & 0.287  \\
    SUDF~\shortcite{paper16} & 0.242  & 0.256  & 0.723  & 0.250  \\
    SMN~\shortcite{paper42} & 0.069  & 0.658  & 0.916  & 0.718  \\
    DMSSN~\shortcite{paper45} & 0.068  & 0.564  & 0.937  & 0.703  \\
    \midrule
    Hyper-HRNet & \textbf{0.056}   & \textbf{0.770}   & \textbf{0.953}      &\textbf{0.810}  \\
    \bottomrule
    \end{tabular}%
  \label{tab:HS-SOD quantitative}%
  \caption{Quantitative Results on the HS-SOD Dataset.}
  \vspace{-2mm}
\end{table}

\noindent\textbf{Efficiency Analysis}. Table \ref{tab:HSOD-BIT quantitative} presents computational complexity of our method, excluding traditional approaches. Unlike RGB-based methods, which often rely on complex network structures and neglect spectral dimensions, our method significantly reduces both parameters and FLOPs. However, the complexity of HRNet results in a larger model size. To improve efficiency, we introduce Hyper-HRNet-Lite with the lightweight Lite-HRNet backbone~\shortcite{paper49}. 
Among HSI-based methods, SUDF uses CNNs for only feature extraction followed by manifold learning and superpixel clustering, leading to low parameters but high FLOPs, while DMSSN reduces parameters via knowledge distillation. Direct parameter comparisons with them are less meaningful. Hyper-HRNet-Lite minimizes FLOPs and achieves comparable performance to the soTA method SMN with fewer parameters, balancing efficiency, speed, and efficacy.

\noindent\textbf{Visualization of Spetral Attention Feature}. Figure \ref{fig:hotmap} shows the Spectral Attention features from our proposed HAR, along with the pseudo-color images and ground truth. These features emphasize the spectral characteristics of HSIs, enhancing the contrast between salient objects and backgrounds while preserving crucial spectral information.

\subsection{Results on HS-SOD Dataset}

\noindent\textbf{Quantitative Analysis}. compares Hyper-HRNet with existing HSI-based methods on the HS-SOD dataset, using consistent training configurations from previous works~\shortcite{paper42,paper45}, utilizing 48 data for training and the rest for testing. Hyper-HRNet outperforms DMSSN and SMN, improving AUC by 0.160 and 0.037, CC by 0.107 and 0.092, and reducing MAE by 0.012 and 0.013, respectively.

\begin{figure}[t]
\centering
\includegraphics[width=\columnwidth]{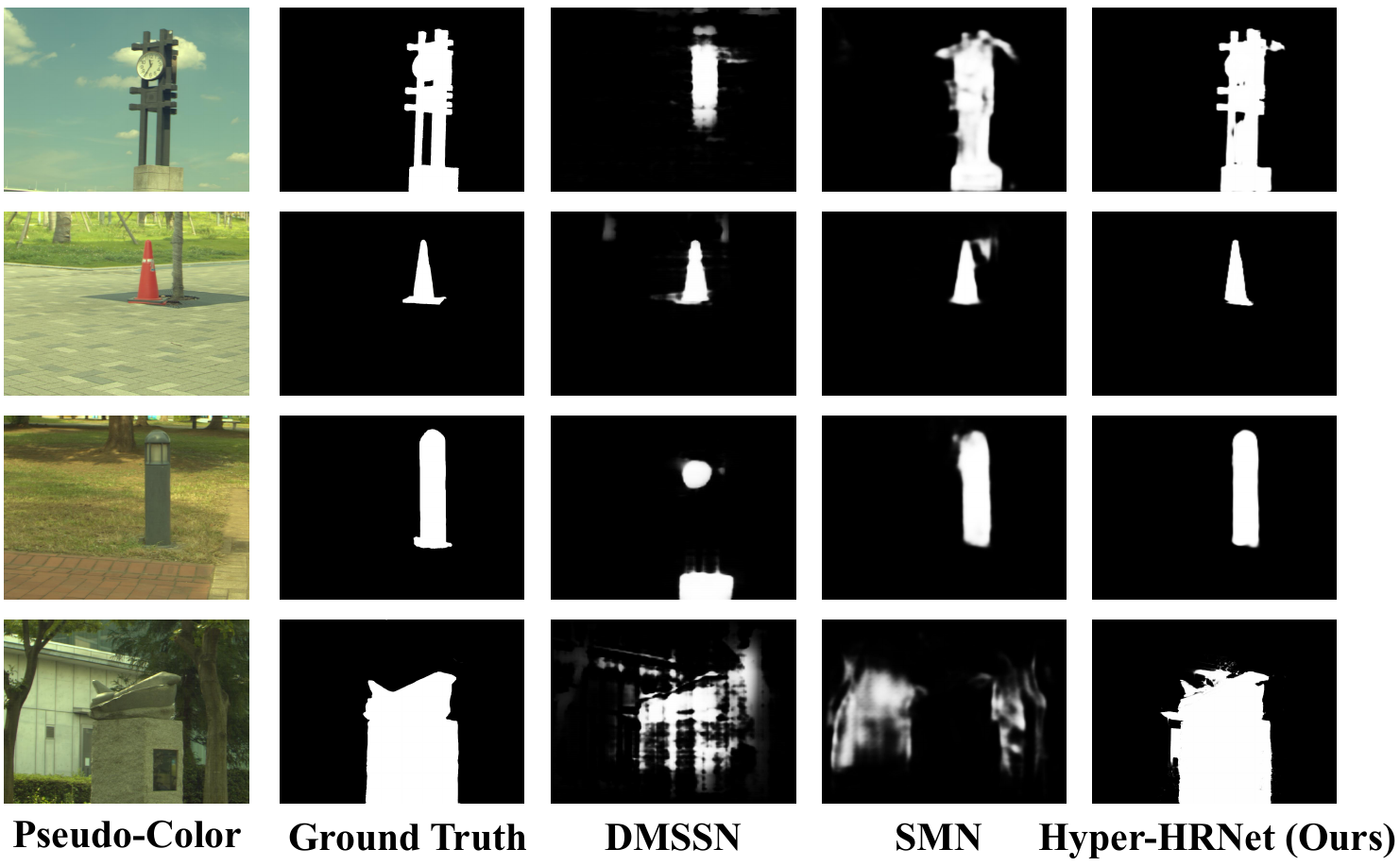}
\caption{Qualitative results on HS-SOD dataset.} 
\label{fig:HS-SOD qualitative}
\end{figure}

\begin{table}[!t]
  \centering
  \renewcommand{\arraystretch}{0.8}
    \setlength{\tabcolsep}{7pt}
    \footnotesize
    \begin{tabular}{cc|cccc}
      \toprule
      HAR  & GTPD  & $MAE\downarrow$  & $avgF_{1} \uparrow$ & $AUC\uparrow$  & $CC\uparrow$ \\
      \midrule
        \ding{51}     & \ding{55}    & 0.034  & 0.508  & 0.883  & 0.655  \\
        \ding{55}    &  \ding{51}     & 0.030  & 0.534  & 0.890  & 0.692  \\
      \midrule
       \ding{51}     &  \ding{51}     & \textbf{0.028} & \textbf{0.589} & \textbf{0.988} & \textbf{0.737} \\
      \bottomrule
    \end{tabular}%
  \caption{Ablation Study of Key Components.}
  \label{tab:Overall ablation}%
  \vspace{-2mm}
\end{table}%

\noindent\textbf{Qualitative Analysis}. Figure \ref{fig:HS-SOD qualitative} presents visual comparisons of Hyper-HRNet against other HSI-based methods on the HS-SOD dataset. While other methods struggle with blurry edges and recognition distortions, Hyper-HRNet leverages reconstructed hyperspectral information to produce saliency maps with clear and precise contours, demonstrating its superior performance in the HSOD task.

\subsection{Ablation Study}
We conducted ablation studies on our HSOD-BIT-V2.

\noindent\textbf{Effect of key componets}. To validate the efficacy of each component within Hyper-HRNet, as detailed in Table \ref{tab:Overall ablation}, we conducted a comparative analysis using HRNet as the baseline. The results indicate substantial performance improvements with the separate integration of HAR and GTPD. Moreover, their combined integration achieves superior results, confirming the effectiveness of the two components.

\begin{table}[tp]
  \centering
  \setlength{\tabcolsep}{0.6mm}
  \renewcommand{\arraystretch}{0.8}
    \footnotesize
    \begin{tabular}{cccc|cccc}
      \toprule
      Interpolate & PCA    & Conv & HAR  & $MAE\downarrow$  & $avgF_{1} \uparrow$ & $AUC\uparrow$  & $CC\uparrow$ \\
      \midrule
       \ding{55}    &  \ding{55}    &  \ding{55}    &  \ding{55}      & 0.169 & 0.178 & 0.664 & 0.193 \\
       \ding{51}     &  \ding{55}    &  \ding{55}    &  \ding{55}      & 0.084  & 0.283  & 0.728  & 0.268  \\
       \ding{55}    &  \ding{51}     &  \ding{55}    &  \ding{55}      & 0.073  & 0.321  & 0.754  & 0.447  \\
       \ding{55}    &  \ding{55}    &  \ding{51}     &  \ding{55}     & 0.079  & 0.426  & 0.833  & 0.472  \\
      \midrule
       \ding{55}    &  \ding{55}    &  \ding{55}     &  \ding{51}    & \textbf{0.028 } & \textbf{0.589 } & \textbf{0.988 } & \textbf{0.737 } \\
      \bottomrule
    \end{tabular}%
  \caption{Comparative Experiments between Different Dimensionality Reduction Methods and HAR.}
  \label{tab:HAR ablation}%
\end{table}%

\begin{table}[tp]
  \centering
  \renewcommand{\arraystretch}{0.8}
    \setlength{\tabcolsep}{7pt}
    \footnotesize
    \begin{tabular}{c|cccc}
    \toprule
    Method &$MAE\downarrow$  & $avgF_{1} \uparrow$ & $AUC\uparrow$  & $CC\uparrow$  \\
    \midrule
    $\textit{w/o}$ MSSA & 0.088  & 0.496  & 0.832  & 0.545  \\
    $\textit{w/o}$ ASAM & 0.095  & 0.478  & 0.822  & 0.539  \\
    ViT & 0.121  & 0.385  & 0.805  & 0.516  \\
    MSST & 0.114  & 0.467  & 0.825  & 0.536  \\
    \midrule
    Hyper-HRNet & \textbf{0.028 } & \textbf{0.589 } & \textbf{0.988 } & \textbf{0.737 } \\
    \bottomrule
    \end{tabular}%
    \caption{Ablation Study of HAR.}
  \label{tab:HAR ablation2}%
\end{table}%

\begin{table}[!htp]
  \centering
  \renewcommand{\arraystretch}{0.8}
    \setlength{\tabcolsep}{7pt}
    \footnotesize
    \begin{tabular}{c|cccc}
    \toprule
    Method &$MAE\downarrow$  & $avgF_{1} \uparrow$ & $AUC\uparrow$  & $CC\uparrow$  \\
    \midrule
    $\textit{w/o}$ CMFI & 0.039  & 0.523  & 0.880  & 0.616  \\
    $\textit{w/o}$ GAFA & 0.044  & 0.522  & 0.899  & 0.617  \\
    $\textit{w/o}$ TAW & 0.039  & 0.526  & 0.920  & 0.666  \\
    \midrule
    Hyper-HRNet & \textbf{0.028 } & \textbf{0.589 } & \textbf{0.988 } & \textbf{0.737 } \\
    \bottomrule
    \end{tabular}%
    \caption{Ablation Study of GTPD.}
  \label{tab:GTPD ablation}%
\end{table}%

\noindent\textbf{Effect of HAR}. To validate the effectiveness of minimizing spectral redundancy in HAR, as shown in Table \ref{tab:HAR ablation}, we conducted comparative experiments on dimensionality reduction methods using Hyper-HRNet without dimensionality reduction as the baseline. The results show significant performance gains with the integration of HAR. Specifically, HAR enhances AUC by 0.050 and CC by 0.183, while reducing MAE by 0.045 compared to the Convolution Layer.

To further validate the effectiveness of HPSA, as shown in Table \ref{tab:HAR ablation2}, we individually removed its key components, labeled as $\textit{w/o}$ MSSA and $\textit{w/o}$ ASAM, and replaced HPSA with other self-attention mechanisms, including the conventional self-attention mechanism in ViT~\shortcite{VIT} and spectral-spatial hybrid attention mechanism MSS~\shortcite{paper45}. Removing MSSA or ASAM consistently led to a performance decline, with both ViT's spatial attention and HSOD's MSSA underperforming relative to HPSA.
These results confirm the effectiveness of HAR and HPSA.

\noindent\textbf{Effect of GTPD}. Table \ref{tab:GTPD ablation} validates the effectiveness of the key modules within GTPD: CMFI, GAFA, and TAW. We conducted experiments by removing these modules individually, labeled as $\textit{w/o}$ CMFI, $\textit{w/o}$ GAFA, and $\textit{w/o}$ TAW. Removing cross-level multi-scale feature interaction, global attention saliency map, or ternary contour-aware weights consistently led to decreased prediction performance. These findings highlight the importance of three critical modules.

\section{Conclusion}
In this work, we introduce HSOD-BIT-V2, the largest and most challenging HSOD dataset to date, and propose a novel high-resolution network, Hyper-HRNet. 
Our dataset includes eight natural backgrounds and five challenging attributes that highlight the spectral advantages of HSIs. 
Our method optimizes HSI utilization, reduces spectral dimensionality, and preserves key spectral information. 
Additionally, it also accurately locates object contours through capturing intact global information and ternary contour-aware saliency. 
While we believe this work will advance HSOD research and establishes a new benchmark for future research, opportunities for improvement remain.
Future efforts will focus on expanding the dataset and advancing hyperspectral image dimensionality reduction and reconstruction.

\clearpage
\setcounter{secnumdepth}{0}

\section{Acknowledgments}
This work was financially supported by the National Key Scientific Instrument and Equipment Development Project of China (No. 61527802), the National Natural Science Foundation of China (No. 62101032), the Young Elite Scientist Sponsorship Program of China Association for Science and Technology (No. YESS20220448), and the Young Elite Scientist Sponsorship Program of Beijing Association for Science and Technology (No. BYESS2022167).

\bibliography{aaai25}

\begin{thebibliography}{31}
\providecommand{\natexlab}[1]{#1}

\bibitem[{Ahmadi, Karimi, and Samavi(2021)}]{paper5}
Ahmadi, M.; Karimi, N.; and Samavi, S. 2021.
\newblock Context-aware saliency detection for image retargeting using convolutional neural networks.
\newblock \emph{Multimedia Tools and Applications}, 80: 11917--11941.

\bibitem[{Borji et~al.(2019)Borji, Cheng, Hou, Jiang, and Li}]{paper21}
Borji, A.; Cheng, M.-M.; Hou, Q.; Jiang, H.; and Li, J. 2019.
\newblock Salient object detection: A survey.
\newblock \emph{Computational visual media}, 5: 117--150.

\bibitem[{Chakrabarti and Zickler(2011)}]{paper37}
Chakrabarti, A.; and Zickler, T. 2011.
\newblock Statistics of real-world hyperspectral images.
\newblock In \emph{CVPR 2011}, 193--200. IEEE.

\bibitem[{Chen et~al.(2021)Chen, Li, Deng, and Lin}]{paper9}
Chen, H.; Li, Y.; Deng, Y.; and Lin, G. 2021.
\newblock CNN-based RGB-D salient object detection: Learn, select, and fuse.
\newblock \emph{International Journal of Computer Vision}, 129(7): 2076--2096.

\bibitem[{Chen et~al.(2024)Chen, Zhao, Xu, Shi, Zhou, Liu, and Li}]{chenhuan}
Chen, H.; Zhao, W.; Xu, T.; Shi, G.; Zhou, S.; Liu, P.; and Li, J. 2024.
\newblock Spectral-Wise Implicit Neural Representation for Hyperspectral Image Reconstruction.
\newblock \emph{IEEE Transactions on Circuits and Systems for Video Technology}, 34(5): 3714--3727.

\bibitem[{Dosovitskiy(2020)}]{VIT}
Dosovitskiy, A. 2020.
\newblock An image is worth 16x16 words: Transformers for image recognition at scale.
\newblock \emph{arXiv preprint arXiv:2010.11929}.

\bibitem[{Foster, Nascimento, and Amano(2004)}]{paper36}
Foster, D.~H.; Nascimento, S.~M.; and Amano, K. 2004.
\newblock Information limits on neural identification of colored surfaces in natural scenes.
\newblock \emph{Visual neuroscience}, 21(3): 331--336.

\bibitem[{Huang et~al.(2021)Huang, Cai, Lin, Zheng, Mao, Qian, Peng, Zhou, Iwamoto, Han et~al.}]{paper2}
Huang, H.; Cai, M.; Lin, L.; Zheng, J.; Mao, X.; Qian, X.; Peng, Z.; Zhou, J.; Iwamoto, Y.; Han, X.-H.; et~al. 2021.
\newblock Graph-based pyramid global context reasoning with a saliency-aware projection for covid-19 lung infections segmentation.
\newblock In \emph{ICASSP 2021-2021 IEEE International Conference on Acoustics, Speech and Signal Processing (ICASSP)}, 1050--1054. IEEE.

\bibitem[{Hughes(2003)}]{paper48}
Hughes, G. 2003.
\newblock On the mean accuracy of statistical pattern recognizers.
\newblock \emph{IEEE Transactions on Information Theory}, 14(1): 55--63.

\bibitem[{{\.I}mamo{\u{g}}lu et~al.(2019){\.I}mamo{\u{g}}lu, Ding, Fang, Kanezaki, Kouyama, and Nakamura}]{paper16}
{\.I}mamo{\u{g}}lu, N.; Ding, G.; Fang, Y.; Kanezaki, A.; Kouyama, T.; and Nakamura, R. 2019.
\newblock Salient object detection on hyperspectral images using features learned from unsupervised segmentation task.
\newblock In \emph{ICASSP 2019-2019 IEEE International Conference on Acoustics, Speech and Signal Processing (ICASSP)}, 2192--2196. IEEE.

\bibitem[{Imamoglu et~al.(2018)Imamoglu, Oishi, Zhang, Ding, Fang, Kouyama, and Nakamura}]{paper17}
Imamoglu, N.; Oishi, Y.; Zhang, X.; Ding, G.; Fang, Y.; Kouyama, T.; and Nakamura, R. 2018.
\newblock Hyperspectral image dataset for benchmarking on salient object detection.
\newblock In \emph{2018 Tenth international conference on quality of multimedia experience (qoMEX)}, 1--3. IEEE.

\bibitem[{Itti, Koch, and Niebur(1998)}]{paper15}
Itti, L.; Koch, C.; and Niebur, E. 1998.
\newblock A model of saliency-based visual attention for rapid scene analysis.
\newblock \emph{IEEE Transactions on pattern analysis and machine intelligence}, 20(11): 1254--1259.

\bibitem[{Le~Moan et~al.(2013)Le~Moan, Mansouri, Hardeberg, and Voisin}]{paper29}
Le~Moan, S.; Mansouri, A.; Hardeberg, J.~Y.; and Voisin, Y. 2013.
\newblock Saliency for spectral image analysis.
\newblock \emph{IEEE Journal of Selected Topics in Applied Earth Observations and Remote Sensing}, 6(6): 2472--2479.

\bibitem[{Li et~al.(2022)Li, Fang, Zha, Gao, and Zheng}]{paper33}
Li, G.; Fang, Q.; Zha, L.; Gao, X.; and Zheng, N. 2022.
\newblock HAM: Hybrid attention module in deep convolutional neural networks for image classification.
\newblock \emph{Pattern Recognition}, 129: 108785.

\bibitem[{Liang et~al.(2013)Liang, Zhou, Bai, and Qian}]{paper28}
Liang, J.; Zhou, J.; Bai, X.; and Qian, Y. 2013.
\newblock Salient object detection in hyperspectral imagery.
\newblock In \emph{2013 IEEE International conference on image processing}, 2393--2397. IEEE.

\bibitem[{Liu et~al.(2023)Liu, Xu, Chen, Zhou, Qin, and Li}]{paper42}
Liu, P.; Xu, T.; Chen, H.; Zhou, S.; Qin, H.; and Li, J. 2023.
\newblock Spectrum-driven Mixed-frequency Network for Hyperspectral Salient Object Detection.
\newblock \emph{IEEE Transactions on Multimedia}.

\bibitem[{Nascimento, Ferreira, and Foster(2002)}]{paper35}
Nascimento, S.~M.; Ferreira, F.~P.; and Foster, D.~H. 2002.
\newblock Statistics of spatial cone-excitation ratios in natural scenes.
\newblock \emph{JOSA A}, 19(8): 1484--1490.

\bibitem[{Peng et~al.(2017)Peng, Zhang, Yu, Luo, and Sun}]{paper39}
Peng, C.; Zhang, X.; Yu, G.; Luo, G.; and Sun, J. 2017.
\newblock Large kernel matters--improve semantic segmentation by global convolutional network.
\newblock In \emph{Proceedings of the IEEE conference on computer vision and pattern recognition}, 4353--4361.

\bibitem[{Qin et~al.(2024)Qin, Xu, Liu, Xu, and Li}]{paper45}
Qin, H.; Xu, T.; Liu, P.; Xu, J.; and Li, J. 2024.
\newblock DMSSN: Distilled Mixed Spectral-Spatial Network for Hyperspectral Salient Object Detection.
\newblock \emph{IEEE Transactions on Geoscience and Remote Sensing}.

\bibitem[{Qin et~al.(2020)Qin, Zhang, Huang, Dehghan, Zaiane, and Jagersand}]{paper43}
Qin, X.; Zhang, Z.; Huang, C.; Dehghan, M.; Zaiane, O.~R.; and Jagersand, M. 2020.
\newblock U2-Net: Going deeper with nested U-structure for salient object detection.
\newblock \emph{Pattern recognition}, 106: 107404.

\bibitem[{Qin et~al.(2019{\natexlab{a}})Qin, Zhang, Huang, Gao, Dehghan, and Jagersand}]{paper25}
Qin, X.; Zhang, Z.; Huang, C.; Gao, C.; Dehghan, M.; and Jagersand, M. 2019{\natexlab{a}}.
\newblock Basnet: Boundary-aware salient object detection.
\newblock In \emph{Proceedings of the IEEE/CVF conference on computer vision and pattern recognition}, 7479--7489.

\bibitem[{Qin et~al.(2019{\natexlab{b}})Qin, Zhang, Huang, Gao, Dehghan, and Jagersand}]{paper44}
Qin, X.; Zhang, Z.; Huang, C.; Gao, C.; Dehghan, M.; and Jagersand, M. 2019{\natexlab{b}}.
\newblock Basnet: Boundary-aware salient object detection.
\newblock In \emph{Proceedings of the IEEE/CVF conference on computer vision and pattern recognition}, 7479--7489.

\bibitem[{Tang et~al.(2021)Tang, Li, Zhong, Ding, and Song}]{paper27}
Tang, L.; Li, B.; Zhong, Y.; Ding, S.; and Song, M. 2021.
\newblock Disentangled high quality salient object detection.
\newblock In \emph{Proceedings of the IEEE/CVF international conference on computer vision}, 3580--3590.

\bibitem[{Wang et~al.(2020)Wang, Sun, Cheng, Jiang, Deng, Zhao, Liu, Mu, Tan, Wang et~al.}]{paper18}
Wang, J.; Sun, K.; Cheng, T.; Jiang, B.; Deng, C.; Zhao, Y.; Liu, D.; Mu, Y.; Tan, M.; Wang, X.; et~al. 2020.
\newblock Deep high-resolution representation learning for visual recognition.
\newblock \emph{IEEE transactions on pattern analysis and machine intelligence}, 43(10): 3349--3364.

\bibitem[{Wang et~al.(2024)Wang, Chen, Li, Xu, Zhao, Duan, Gao, and Lin}]{wangze}
Wang, Z.; Chen, H.; Li, J.; Xu, T.; Zhao, Z.; Duan, Z.; Gao, S.; and Lin, X. 2024.
\newblock Opto-intelligence spectrometer using diffractive neural networks.
\newblock \emph{Nanophotonics}, 13(20): 3883--3893.

\bibitem[{Wu et~al.(2022)Wu, Su, Tao, Han, Paoletti, Haut, Plaza, and Plaza}]{paper14}
Wu, Z.; Su, H.; Tao, X.; Han, L.; Paoletti, M.~E.; Haut, J.~M.; Plaza, J.; and Plaza, A. 2022.
\newblock Hyperspectral anomaly detection with relaxed collaborative representation.
\newblock \emph{IEEE Transactions on Geoscience and Remote Sensing}, 60: 1--17.

\bibitem[{Xie et~al.(2017)Xie, Girshick, Doll{\'a}r, Tu, and He}]{paper38}
Xie, S.; Girshick, R.; Doll{\'a}r, P.; Tu, Z.; and He, K. 2017.
\newblock Aggregated residual transformations for deep neural networks.
\newblock In \emph{Proceedings of the IEEE conference on computer vision and pattern recognition}, 1492--1500.

\bibitem[{Yu et~al.(2021)Yu, Xiao, Gao, Yuan, Zhang, Sang, and Wang}]{paper49}
Yu, C.; Xiao, B.; Gao, C.; Yuan, L.; Zhang, L.; Sang, N.; and Wang, J. 2021.
\newblock Lite-HRNet: A Lightweight High-Resolution Network.
\newblock In \emph{CVPR}.

\bibitem[{Yun and Lin(2023)}]{paper26}
Yun, Y.~K.; and Lin, W. 2023.
\newblock Towards a complete and detail-preserved salient object detection.
\newblock \emph{IEEE Transactions on Multimedia}.

\bibitem[{Zhang et~al.(2018)Zhang, Zhang, Yan, Gao, and Wei}]{paper30}
Zhang, L.; Zhang, Y.; Yan, H.; Gao, Y.; and Wei, W. 2018.
\newblock Salient object detection in hyperspectral imagery using multi-scale spectral-spatial gradient.
\newblock \emph{Neurocomputing}, 291: 215--225.

\bibitem[{Zhao et~al.(2015)Zhao, Ouyang, Li, and Wang}]{paper24}
Zhao, R.; Ouyang, W.; Li, H.; and Wang, X. 2015.
\newblock Saliency detection by multi-context deep learning.
\newblock In \emph{Proceedings of the IEEE conference on computer vision and pattern recognition}, 1265--1274.

\end{thebibliography}

\end{document}